\documentclass[letterpaper, 10 pt, conference]{ieeeconf}  

\IEEEoverridecommandlockouts                              

\title{\LARGE \bf
Spectral Temporal Graph Neural Network for Trajectory Prediction
}

\author{Defu Cao*,
        Jiachen Li*,
        Hengbo Ma,
        Masayoshi Tomizuka,~\IEEEmembership{Life~Fellow,~IEEE}
\thanks{*The authors contributed equally to this work.}
\thanks{D. Cao is with Peking University, China (Email: cdf@pku.edu.cn).}
\thanks{J. Li, H. Ma and M. Tomizuka are with University of California, Berkeley, USA (Email: \{jiachen\_li, hengbo\_ma, tomizuka\}@berkeley.edu).}

}

\usepackage{subfigure}
\usepackage{graphicx}
\usepackage{comment}
\usepackage{amsmath,amssymb} 
\usepackage{color}
\usepackage{bm}
\usepackage{amsfonts}
\usepackage{array}
\usepackage{algorithm}
\usepackage{algorithmic}
\usepackage{cite}
\usepackage{booktabs}
\usepackage{float}
\usepackage{multirow}
\usepackage{mathrsfs}
\usepackage{amsfonts,amssymb} 
\usepackage{lipsum}
\usepackage{wrapfig}
\usepackage{caption}
\usepackage{color}

\begin{document}

\maketitle

\begin{abstract}
An effective understanding of the contextual environment and accurate motion forecasting of surrounding agents is crucial for the development of autonomous vehicles and social mobile robots. 
This task is challenging since the behavior of an autonomous agent is not only affected by its own intention, but also by the static environment and surrounding dynamically interacting agents. 
Previous works focused on utilizing the spatial and temporal information in time domain while not sufficiently taking advantage of the cues in frequency domain.
To this end, we propose a Spectral Temporal Graph Neural Network (SpecTGNN), which can capture inter-agent correlations and temporal dependency simultaneously in frequency domain in addition to time domain. 
SpecTGNN operates on both an agent graph with dynamic state information and an environment graph with the features extracted from context images in two streams. 
The model integrates graph Fourier transform, spectral graph convolution and temporal gated convolution to encode history information and forecast future trajectories.
Moreover, we incorporate a multi-head spatio-temporal attention mechanism to mitigate the effect of error propagation in a long time horizon.
We demonstrate the performance of SpecTGNN on two public trajectory prediction benchmark datasets, which achieves state-of-the-art performance in terms of prediction accuracy.

\end{abstract}

\IEEEpeerreviewmaketitle

\section{Introduction}

Understanding the behavior of traffic participants is essential for autonomous driving systems, especially for safe navigation in crowded and complex traffic scenarios \cite{lefevre2014survey}. 
In these scenarios, accurate trajectory prediction can not only help autonomous vehicles make informed decisions and safely plan their future motions \cite{li2020interaction}, but also help surveillance systems detect abnormal behaviors . 
For short-term prediction, using purely physical models or traditional statistical models can obtain acceptable performance since the behavior is unlikely to be largely affected by external factors in a short period (i.e., less than a second). 
However, these approaches are not sufficient for more complicated long-term prediction.
Given a series of historical observations, there may be a variety of plausible future motions due to different human intentions or mutual influence between agents \cite{zhao2019multi,li2020evolvegraph}. 
The inherent uncertainty in forecasting the future makes long-term trajectory prediction a challenging task. 
Besides, the prediction system is also expected to discriminate traversable areas delineated by the road boundaries and the right of way in accordance with the traffic rules \cite{hong2019rules}.
Therefore, it is necessary to design effective modules to model the interaction between interactive agents as well as to exploit the environmental context information.

Many recent works attempt to model the interaction between traffic participants by constructing a scene graph and utilizing graph neural networks to extract relational features \cite{yan2018spatial}. 
However, previous methods did not explicitly utilize the information in frequency domain. 
Inspired by \cite{cao2020spectral}, for time series signals whose primary information content lies in localized singularities, frequency domain can provide a much more compact representation than the original time domain.
In this work, we propose to additionally exploit the patterns extracted in frequency domain and design a SpecTGNN unit that employs a new graph convolution method based on spectral decomposition to capture the information from different frequency components.
The proposed method can simultaneously process temporal correlations and spatial structures. 
More specifically, we take advantage of the graph Fourier transform (GFT) and temporal convolution to model multi-agent trajectories in frequency domain.
The underlying intuition is that the spectral representation obtained by GFT may present patterns that can be utilized more effectively by graph convolution operations.
After obtaining the intermediate output of the SpecTGNN unit, we also apply a multi-head spatio-temporal attention to figure out relative importance of the information of each agent at each time step, which can reduce the effect of error propagation in long-term prediction \cite{2020GMAN,li2019conditional}.

Moreover, most existing methods use the state information of agents to build the scene graph while the environment information only serves as additional features, which might not be fully exploited by the model. 
To address this issue, we construct an environment graph in addition to the agent graph with the image features extracted by a convolutional neural network (CNN) followed by a fully connected (FC) network.
The features extracted from context images serve as a representation of abstract relationship between the agents, since the environment may also influence the interactive behaviors. 
These features are used to construct the Laplacian matrix of the environment graph, enabling the spectral graph convolution on the environment information.

To the best of our knowledge, we are the first to utilize spatial correlations and temporal dependency simultaneously in frequency domain to handle the prediction task. 
The main contributions are summarized as follows:
\begin{itemize}
    \item We propose a Spectral Temporal Graph Neural Network (SpecTGNN) for multi-agent trajectory prediction.
    SpecTGNN integrates the advantages of sequence modeling with convolutional networks and feature extraction in frequency domain via graph Fourier transform and graph convolution operations in a unified framework. 
    \item We propose a SpecTGNN unit which consists of two blocks particularly used to extract information on the agent graph and environment graph, respectively. The SpecTGNN unit is designed to extract trajectory patterns and capturing dynamic frequency correlations. 
    \item We validate SpecTGNN on two prediction benchmark datasets, which achieves state-of-the-art performance. 
\end{itemize}

\section{Related Work}

\subsection{Behavior and Trajectory Prediction}
Traditionally, trajectory prediction problems are widely studied in statistics, signal processing, and systems engineering. 
However, these traditional methods heavily rely on prior knowledge which may not be available in the data. 
Therefore, more advanced data-driven methods such as deep learning have been extensively explored for solving trajectory prediction problems.
In recent years, many approaches have been introduced to model the trajectory information, such as sequence modeling based on recurrent neural networks (RNNs) \cite{lee2017desire,fernando2018soft+,ma2019wasserstein}.
In order to model the mutual influence between agents, researchers proposed various techniques to aggregate the information, such as social pooling~\cite{2016Social}, attention mechanisms~\cite{fernando2018soft+}, and graph message passing~\cite{mohamed2020social}. 
Mixture models are also employed to encourage the multi-modality and enable different outcomes in trajectory prediction \cite{cui2019multimodal,zyner2019naturalistic,zhan2018towards}.
Besides, deep generative models are utilized to enhance the performance of distribution learning to generate better prediction hypotheses \cite{sadeghian2019sophie,zhao2019multi,bhattacharyya2019conditional,li2019interaction}. 
Unlike most existing works that only utilize the spatial and temporal information in time domain, we propose to also leverage the patterns extracted in frequency domain.

\subsection{Graph Neural Networks}

Graph neural networks (GNN) have achieved outstanding performance on different types of tasks in various domains. Many variants of the model architecture and message passing rules have been proposed, such as GCN \cite{kipf2016semi}, Spectral GCN \cite{bruna2013spectral}, Spatial GCN \cite{bruna2013spectral}, ChebNet~\cite{Defferrard2016Convolutional}, GraphSAGE~\cite{hamilton2017inductive,ma2020reinforcement} and GAT~\cite{velivckovic2017graph}.
In recent years, researchers attempt to leverage GNN to incorporate relational inductive biases in the learning based models to solve various real-world tasks such as traffic flow forecasting and trajectory prediction \cite{cao2020spectral,zhao2020multivariate, li2017diffusion,yu2017spatio,zhang2019stochastic,mohamed2020social,choi2020shared}. 
Li et al proposed a deep learning framework for traffic forecasting that incorporates both spatial and temporal dependency in the traffic flow \cite{li2017diffusion}. 
Yu et al integrated graph convolution and gated temporal convolution through spatio-temporal convolutional blocks for traffic prediction \cite{yu2017spatio}. 
Besides, Zhang et al used a social graph with a hierarchical LSTMs to solve the challenge of complexity of real-world human social behaviors and uncertainty of the future motion \cite{zhang2019stochastic}. 
Mohamed et al presented a kernel function to embed the social interaction between pedestrians in the adjacency matrix \cite{mohamed2020social}. 
For a more comprehensive literature review on graph neural network, please refer to recent surveys~\cite{wu2020comprehensive, zhou2018graph, zhang2020deep}.
In this work, we take advantage of the graph convolution to extract patterns in frequency domain to model the agents' interactions and the environment.

\section{Problem Formulation}

The goal of this work is to predict the future trajectories of multiple interactive agents based on their historical states and the context information. 
Without lose of generality, we assume that $N$ agents are navigating within the observation area. The number of involved agents can be flexible in different situations. The involved agents include vehicles, pedestrians and cyclists. 
The history horizon and the forecasting horizon are denoted as $T_h$ and $T_f$, respectively.  
We denote the whole trajectories of $N$ agents as:
\begin{equation}
\small
  \begin{split}
  \bm{T}_{1:T} = \{\tau^i_{1:T}| \tau^{i}_t = (x_t^i,y_t^i), T = T_h+T_f,i=1,...,N\},
  \end{split}
  \end{equation}
where $(x_t^i,y_t^i)$ is the position of agent $i$ at time $t$. 
The coordinates can be either in the world space or the image pixel space. 
The goal of prediction is to approximate the conditional distribution of future trajectories given the history observations. 

\section{Method: SpecTGNN}

We provide a brief overview of the proposed method.
First, we build an agent graph $G^a$ to model the interactive behavior of agents, and an environment graph $G^e$ to encode the context information.
Next, we present a novel SpecTGNN unit which consists of two types of blocks that operate on $G^a$ and $G^e$, respectively. 
The two blocks share the same structure and operations but learn different parameters. They apply spectral temporal graph convolution to the agent state information and environment information to capture the spatial correlations and temporal dependency.
Finally, the intermediate representation in frequency domain obtained by graph convolution can be used to predict the future trajectory with a multi-head spatio-temporal attention mechanism and a set of stacked temporal convolutions with residual connection.

\begin{figure*}
\centering
\includegraphics[width=\textwidth]{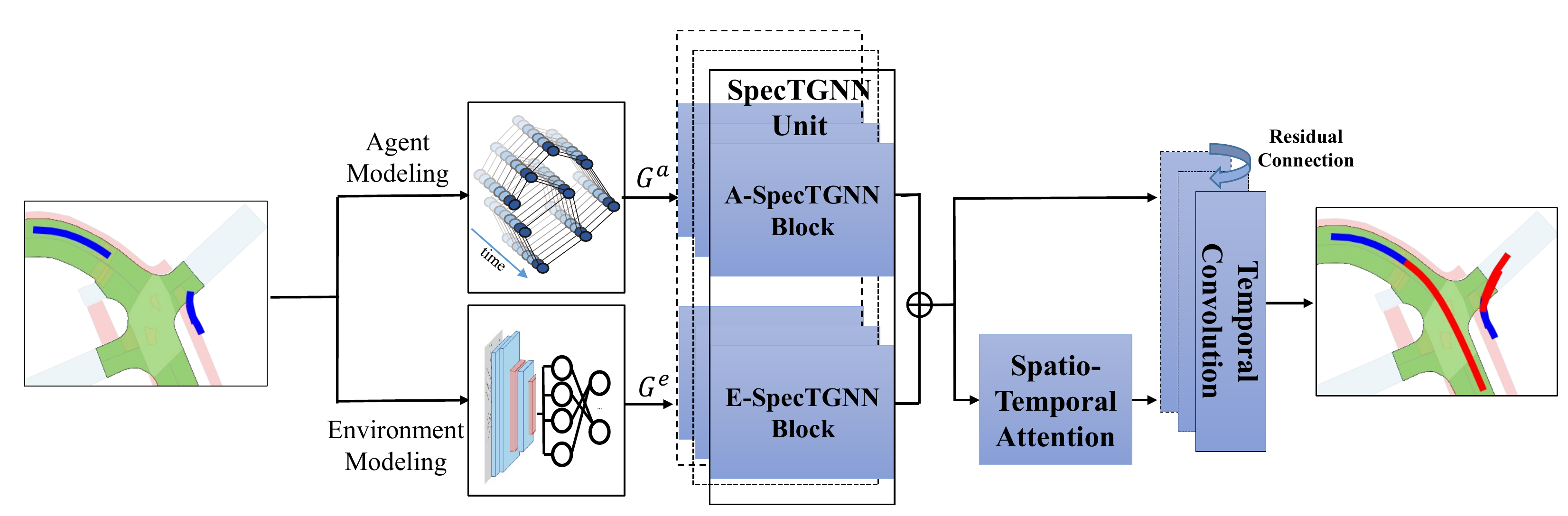}
\caption{The overall architecture of SpecTGNN.
The whole framework consists of three modules: spectral temporal graph convolution (SpecTGNN) unit, spatio-temporal attention mechanism (STAtt) and temporal convolutional neural network (TCNN). 
More specifically, each SpecTGNN unit has two types of SpecTGNN blocks: 1) A-SpecTGNN block to model the agent information 
on the agent graph; and 2) E-SpecTGNN block to model the environment information on the environment graph.
The SpecTGNN unit extracts the spectral and temporal state information of multiple interactive agents and environment information via a spectral graph convolution and a temporal convolution. 
The key operation of STAtt module is to dynamically assign attention weights to different node (agent) attributes at different time steps.
The TCNN module further aggregates the attention weights and node attributes to generate future trajectories. 
}
\vspace{-0.5cm}
\label{fig:network}
\end{figure*}

\subsection{Spectral Temporal Graph Convolutional Unit}
We introduce a novel spectral temporal graph convolutional (SepcTGNN) unit including two types of blocks, which operates on an agent graph $G^a$ and an environment graph $G^e$, respectively.  
As shown in Fig. \ref{fig:network}, the two SpecTGNN blocks share the same model structure but they are applied to different topologies of $G^a$ and $G^e$. 
Multiple layers of the SpecTGNN units with different learnable parameters can be stacked to enhance the model capability of understanding the history information. 

\subsubsection{Agent Modeling}
In order to capture the behavior patterns of a group of interactive agents, we use a graph structure to represent the relationship between different entities, and embed the agent information into node attributes.
The nodes (agents) are fully connected to each other by a set of weighted edges. 

More formally, given the observed trajectories $\bm{T}^{1:T_h}$, we build a spatio-temporal graph $G^{a}_t=(V^a_t,E^a_t)$ with $N$ nodes to represent the information of agent interactions at time $t$. 
$E^{a}_t$ is a set of edges representing the connectivity between the agents in the form of weight matrix.
We have $E^{a}_t=\{w_{ij,t}^{a}|i,j=1,...,N\}$, which depends on the relative distances between agents in $G^a_t$. 
If there is no edge between a certain pair of nodes, the corresponding weight is set to 0.
More formally, the weight matrix $E^{a}_t$ is defined as:
\begin{align}
\begin{split}
 w_{ij,t}^{a}= \left \{
 \begin{array}{ll}
     1 / ||\tau^i_t - \tau^j_t||_2, & i\neq j,\\
     0, & \text{otherwise}.
\end{array}
\right.
\end{split}
\end{align}

\subsubsection{Environment Modeling}

We also use a graph structure $G^e=(V^e,E^e)$ to represent the environment information.
The Laplacian matrix of $G^e$ is built based on the features extracted from
the context image $I$. 
We choose to utilize the VGG backbone ~\cite{simonyan2015very} to extract and encode image features due to its effectiveness on capturing information in images. 
After that, we apply a fully connected layer after the last convolutional layer to yield the edge weight matrix $E^{e}=\{w_{ij}^{e}|i,j=1,...,N\} \in \mathbb{R}^{N \times N}$, where the output feature of the last layer maps to each agent in the environment topology.

\subsubsection{Graph Fourier Transform (GFT) and Inverse Graph Fourier Transform (IGFT)}
Before we employ the spectral graph convolution, 
we apply the convolution operation on graph structures in the spectral domain with graph Fourier transforms.
We introduce the graph convolution operator $*\mathcal{G}$ from the perspective of graph signal processing. 
Note that there are two types of normalized graph Laplacian: agent Laplacian $L^a$ and environment Laplacian $L^e$. 
To simplify notations, we use $*$ to donate the two types.

We use the normalized graph Laplacian $L^*$ to calculate the Fourier basis $U\in \mathbb{R}^{N\times N}$:
\begin{equation}
 L^* = I_N - {D^*}^{-\frac{1}{2}}E^*{D^*}^{-\frac{1}{2}} = U^*\Lambda^* {U^*}^\top,
\end{equation}
where $I_N$ is an identity matrix, $D^*\in \mathbb{R}^{N\times N}$ is the diagonal degree matrix with $D^*_{ii}=\sum_j {E_{ij}^*}^\top$, $U^*$ is the eigenvector matrix of the normalized graph Laplacian $L^*$ sorted by the eigenvalues in the descending order, and $\Lambda^*$ is the diagonalized matrix of the eigenvalues defined as $\Lambda^*_{ii}=\lambda_i$. 
$U^*$ forms an orthogonal space with ${U^*}^\top$: ${U^*}^\top U^*=I$. 
The input of the spatio-temporal graph is composed of the spatial states of $T_h$ time steps, where the used state can be stored as the form of matrix $V^{*} \in \mathbb{R}^{T_h \times N\times c_{\text{in}}}$, which can represent the node information of the agent graph or the environment graph, where $N$ is the number of agents and $c_\text{in}$ is attribute dimension.
The Fourier transform $\mathcal{F}$ and the inverse transform $\mathcal{F}^{-1}$ of input $V^{*}$ in the spectral domain can be defined as:
\begin{equation}
\begin{aligned}
\mathcal{F}^*(V^{*}) ={U^*}^\top V^{*}, \quad
{\mathcal{F}^*}^{-1}(\hat{V}^{*}) = U^* \hat{V}^{*}.
\end{aligned}
\end{equation}
As is defined above, the graph Fourier transform projects the graph signal input $V^{*}$ into the eigenspace constituted by the eigenvectors (rows of $U^*$) of the matrix $L^*$.

\subsubsection{Spectral Graph Convolution (SGConv)}

To simplify notations, we denote the kernel of SGConv operation as $g_\theta$. For 2D graph signal $v \in  \mathbb{R}^{T_h \times N}$,
the graph convolution in frequency domain can be written as:
\begin{equation}
y^*=g_\theta*\mathcal{G}(v) = g_\theta(\Lambda^* {U^*}^\top )v=g_\theta(\Lambda){U^*}^\top v.
\end{equation}
Without loss of generality, we extend the convolution operator $*\mathcal{G}$ to multi-dimensional tensors. For $T_h$ 3D graph signals $V^{*}\in \mathbb{R}^{T_h\times N \times c_\text{in}}$ which has $c_\text{in}$ channels, the 3D graph convolution can be defined as :
\begin{equation}
Y^*_{SG}=g_{*\Theta} U^{*\top} V^{*} = \sum\limits_{k=1}^{c^*_\text{out}}\sum\limits_{j=1}^{c^*_\text{in}} \sum\limits_{i=1}^{N}\Theta ^*_{kij}\Lambda^* U^{*\top} V^{*}, 
\end{equation}
where the graph convolution kernel on 3D variables is denoted as $\Theta^* \in \mathbb{R}^{T_h\times N \times c_{\text{in}}^* \times c^*_\text{out}}$.

\subsubsection{Temporal Gated Convolution (TGConv)}
The agent graph $G^a$ and environment graph $G^e$ are spatio-temporal graphs.
We propose a temporal gated convolutional network in frequency domain, which has the advantages of simple structure and fast training. 
In addition, the information at different time steps in time domain will have no temporal dependence after it is transformed into frequency domain, thus the TGConv can learn the long-interval patterns in sequences effectively.
In the graph Fourier space, the model can capture the global information of history trajectory by mapping the eigenvectors. 
Specifically, we use 1D convolutional method to capture the features of temporal sequences. 
Given the input $\mathcal{F}^*(V^{*})$, the convolution kernel $\mathcal{K}^*\in \mathbb{R}^{1 \times L \times c_{\text{in}}^*\times c_{\text{out}}^*}$ can be represented as the mapping between the input $\mathcal{F}^*(V^{*})$ and the output of temporal gated convolution, where $L$ is the kernel size of $\mathcal{K}^*$, $c_{\text{in}}^*$ and $c_{\text{out}}^*$  represent the input channel and output channel, respectively. 
The formal representation of TGConv can be written as: 
\begin{equation}
Y_{TG}^* = \
\text{TGConv}^*_\mathcal{K}(\mathcal{F}^*(V^{*})). 
\end{equation}

\begin{figure}[!tbp]
\centering
\includegraphics[width=6.cm]{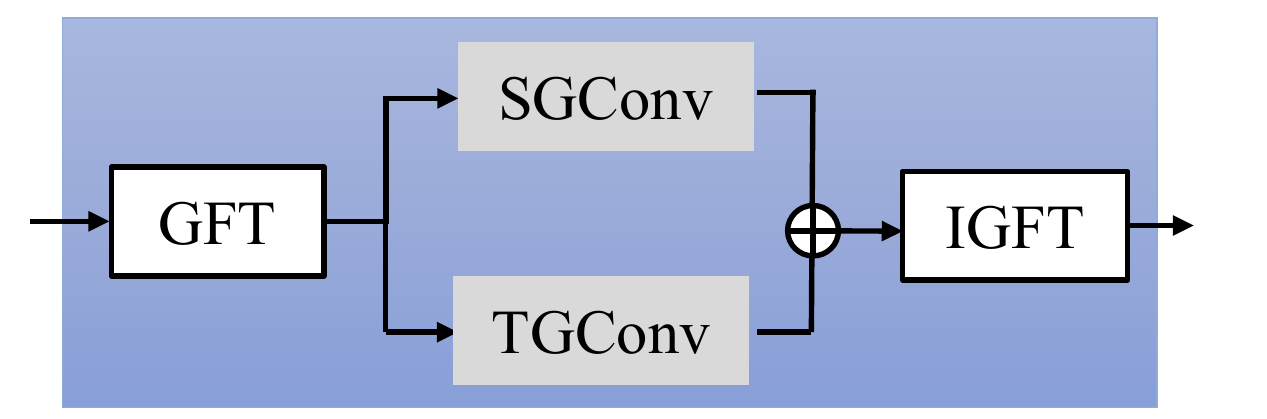}
\caption{An illustrative diagram of the SpecTGNN block.}
\vspace{-0.3cm}
\label{fig:network_block}
\end{figure}

\subsubsection{Spectral Temporal Graph Convolution}
As shown in Fig.~\ref{fig:network_block}, the graph convolution directly processes the graph data and extracts the node attributes based on the history trajectory and the spatial information in the neighborhood. We combine the results of SGConv and TGConv in frequency domain and then pass the embedding through IGFT to yield the embedding vector:
\begin{align}
& Y^*=\mathcal{F^*}^{-1}(Y_{SG}^*+Y_{TG}^*), \nonumber\\
&Y =Y^a + Y^e, Y \in \mathbb{R}^{T_h \times N \times c_\text{out}}. 
\end{align}

\subsection{Spatio-Temporal Attention Mechanism}
\begin{figure}[!tbp]
\centering
\includegraphics[width=0.9\columnwidth]{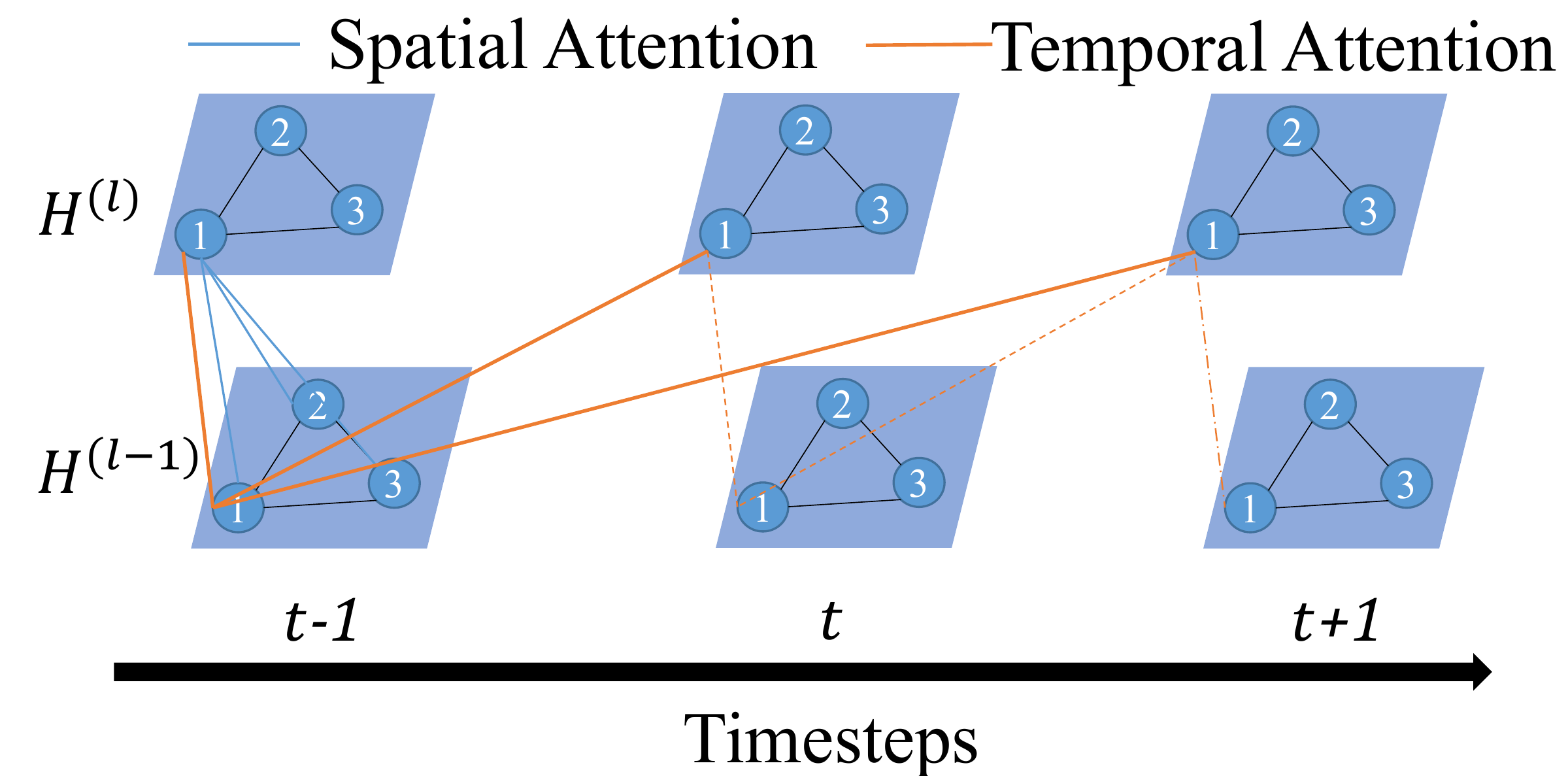}
\caption{An illustrative diagram of the spatio-temporal attention mechanism: $H^{(l)}$ donates the $l$th channel state of each input where $l \in [1,c_\text{in}]$. We can calculate the spatial attention scores by utilizing complete agent information in different channels and calculate the temporal attention scores by using the information of each agent at history time steps. }
\vspace{-0.3cm}
\label{fig:network_attention}
\end{figure}

We design a spatio-temporal attention mechanism to reduce the error propagation effect between different prediction time steps over a long time horizon.
As shown in Fig.~\ref{fig:network_attention}, we apply the spatial attention scores to the spatial interactions of all agents and the temporal attention scores to different time steps to extract spatio-temporal features. 
More specifically, we use the multi-head attention to calculate the attention scores on the spectral embedding given by the SpecTGNN unit.
Formally, the multi-head attention mechanism can be written as:
\begin{align}
& \text{Attention}(\mathbf{Q},\mathbf{K},\mathbf{V}) = \text{Softmax}(\frac{\mathbf{QK}^T}{\sqrt{d_k}})\mathbf{V}, \nonumber\\
& Y_{ST} = \text{MultiHead}(\mathbf{Q},\mathbf{K},\mathbf{V}) = \text{Concat}(\text{head}_i)\mathbf{W}^H, \nonumber \\
& \text{head}_i = \text{Attention}(Y\mathbf{W}_i^Q,Y\mathbf{W}_i^K,Y\mathbf{W}_i^V),
\end{align}
where $d_k$ is the dimension of each head, $\mathbf{Q},\mathbf{K}$ are the query and key, which can be calculated by linear projections with parameters $\mathbf{W}^Q, \mathbf{W}^K \in \mathbb{R}^{c_\text{out}\times d_k}$ respectively. $\mathbf{V}$ is calculated by the spectral embedding ($Y$) given by the final SpecTGNN unit and $\mathbf{W}^V \in  \mathbb{R}^{c_\text{out}\times d_\text{out}}$. $\mathbf{W}^H \in \mathbb{R}^{(N_h\times d_\text{out})\times (N_h\times d_\text{out})}$ projects the concatenation of the $N_h$ head output (each in $\mathbb{R}^{d_\text{out}}$) to the output space $\mathbb{R}^{N_h \times d_\text{out}}$.

\begin{table*}[htbp]
	\centering
	\caption{$\text{minADE}_{20}$ / $\text{minFDE}_{20}$ (pixels) comparisons on the SDD dataset.}
	\vspace{-0.2cm}
	\setlength{\tabcolsep}{1.5mm}{
        \resizebox{\textwidth}{!}{
	\begin{tabular}{l|cccc|ccccc}
		\toprule
		\multirow{2}{*}{Methods} & \multicolumn{4}{c|}{Baseline Methods}&\multicolumn{5}{c}{Proposed Method}
		\\\cmidrule(r){2-10}
		  & CF-VAE~\cite{bhattacharyya2019conditional} & P2TIRL~\cite{deo2020trajectory} & SimAug~\cite{2020SimAug} & PECNet~\cite{mangalam2020not} &\textbf{Base} &\textbf{+TGConv}   & \textbf{+Image}&\textbf{+STAtt} &\textbf{SpecTGNN}\\
		\midrule
		minADE$_{20}$ &   12.60   & 12.58  & 10.27 & 9.96 &9.14 & 8.29  &   8.46&8.65&\textbf{8.21}
    \\
        minFDE$_{20}$  &22.30   & 22.07  & 19.71 & 15.88  
        &13.90&13.60 & 12.57 & 13.36 & \textbf{12.41}
    \\
		\bottomrule
	\end{tabular}}}
	\label{scores_SDD}
\end{table*}

\begin{table*}[htbp]
	\centering
	\caption{$\text{minFDE}_{20}$ (meters) comparisons on the nuScenes (vehicle) dataset.}
	\vspace{-0.2cm}
	\setlength{\tabcolsep}{1.5mm}{
        \resizebox{\textwidth}{!}{
	\begin{tabular}{l|cccccc|ccc}
		\toprule
	\multirow{2}{*}{Time} & \multicolumn{6}{c|}{Baseline Methods}&\multicolumn{3}{c}{Proposed Method}
		\\\cmidrule(r){2-10}
		& CVM  & CSP~\cite{deo2018multi} & CAR-Net~\cite{sadeghian2018car}& SpAGNN~\cite{casas2020spagnn} & STGAT \cite{huang2019stgat} & S-STGCNN~\cite{mohamed2020social}& \textbf{+Image} & \textbf{+STAtt} &\textbf{SpecTGNN} 
		\\
		\midrule
1.0s & 0.32    &0.46  & 0.38   &0.36 & 0.30 &0.35  &0.29 & 0.31   &\textbf{0.28}     \\
2.0s & 0.89   & --  & --   &-- & 0.78 &0.81 &0.77&0.74&\textbf{0.72} \\
3.0s & 1.70  &1.50  & 1.35  &1.23& 1.26 &1.23& 1.21    &1.20&\textbf{1.19}\\
4.0s   & 2.73   & -- &--  &-- & 2.09 &2.15 &1.96&1.92&\textbf{1.87}\\
		\bottomrule
	\end{tabular}}}
	\label{scores_nu}
	\vspace{-0.5cm}
\end{table*}

\subsection{Temporal Convolution Neural Network (TCNN)}

TCNN operates directly on the feature representations from the final SpecTGNN unit $Y$ and STAtt $Y_{ST}$ and expands them as a necessity for prediction~\cite{mohamed2020social}. Excepting for the first TCNN layer, the remaining TCNNs are composed of CNNs with different kernel sizes in a series of residual connections.

\subsection{Loss function}

We assume that the ground truth of predicted trajectory $\tau_t = (x_t, y_t)$ follows a bivariate Gaussian distribution denoted as $\tau_t \sim \mathcal{N}(\bm{\mu}_t,\bm{\sigma}_t)$, 
where $\bm{\mu}_t$, $\bm{\sigma}_t$ are the mean and variance of the distribution. 
Our loss function contains two parts: a likelihood based loss and an distance based loss.
The former is defined as
\begin{equation}
L_{\text{prob}} = -\sum_{t=T_h+1}^{T_f+T_h}\log(\mathbb{P}(\tau_t|\hat{\bm{\mu}}_t,\hat{\bm{\sigma}}_t)),
\end{equation}
where $\hat{\bm{\mu}}_t,\hat{\bm{\sigma}}_t$ are the predicted parameters of the Gaussian distribution. This loss term aims to maximize the log-likelihood of the ground truth. 
The latter is defined as
\begin{equation}
L_{\text{dist}}=\frac{1}{T_f}\sum_{t=T_h+1}^{T_f+T_h}(\tau_t-\hat{\tau}_t)^2,
\end{equation}
which aims to minimize the $L_2$ distance between the predicted trajectory and the ground truth.
The complete loss function is a linear combination of the two parts, which can be written as $L_{\text{total}} =  L_{\text{prob}} + \lambda L_{\text{dist}}$,
where $\lambda$ is a hyperparameter. 
In this paper, we set $\lambda=1$ for all the experiments on both benchmark datasets.

\section{Experiments}

\subsection{Datasets}
\textbf{Stanford Drone Dataset (SDD)}: 
SDD is a large-scale dataset collected in urban scenes in a university campus, which contains images, videos and trajectory annotations of various types of agents such as pedestrians, bicycles and vehicles. 
This dataset includes complex scenarios involving various types of human interactions. 
We adopted the dataset provided in the TrajNet benchmark \cite{becker2018evaluation} to generate our training, validation and test sets. We predicted the future 4.8s (12 frames) based on 3.2s (8 frames) history information.

\textbf{nuScenes Dataset}: 
The nuScenes dataset is a public large-scale dataset for autonomous driving, which consists of 1,000 diverse challenging traffic scenes. Both the trajectory  and map information are provided. Each scene has a length of 20 seconds with a frame rate of 2Hz. We split the training, validation and test sets with a ratio of 60\%, 20\% and 20\%.

\subsection{Baseline and Evaluation Metrics}
We compare SpecTGNN with several state-of-the-art baselines on both datasets, including CF-VAE~\cite{bhattacharyya2019conditional}, P2TIRL~\cite{deo2020trajectory}, SimAug~\cite{2020SimAug}, PECNet~\cite{mangalam2020not}, CSP~\cite{deo2018multi}, CAR-Net~\cite{sadeghian2018car}, SpAGNN~\cite{casas2020spagnn}, STGAT \cite{huang2019stgat}, Social-STGCNN (S-STGCNN)~\cite{mohamed2020social}.

We employ the widely used evaluation metrics for trajectory prediction tasks: minimum average displacement error (minADE) and minimum final displacement error (minFDE). 
More specifically, minADE$_{20}$ is the minimum average $l_2$ distance between the 20 prediction hypotheses and the ground truth over all the involved entities within the prediction horizon. 
minFDE$_{20}$ is the minimum $l_2$ distance between the 20 prediction hypotheses and ground truth at the last time step. 
More formally, we have
\begin{equation}
\small
\begin{aligned}
& \text{minADE}_{20} = \frac{\sum\limits_{n=1}^{N}\min\limits_{k}\sum\limits_{t=T_h+1}^{T_h+T_f}||\hat{\tau}_{t,k}^n - \tau_{t,k}^n||_2}{N \times T_f}, \ k\in\{1,...,20\},\\
& \text{minFDE}_{20} = \frac{\sum\limits_{n=1}^{N}\min\limits_{k}||\hat{\tau}_{T_h+T_f}^n - \tau_{T_h+T_f}^n||_2}{N},  \ k\in\{1,...,20\}.
\end{aligned}
\end{equation}

\subsection{Implementation Details}

SpecTGNN consists of three main components: a SpecTGNN unit, a multi-head spatio-temporal attention mechanism, and a temporal convolution. 
For the SpecTGNN unit, we construct two types of graphs based on the association of agents and the original image. 
The number of nodes in those graphs is equal to the number of agents in the observed area. 
We set a training batch size of 64 on SDD dataset and 128 on nuScences dataset.
We train the model on a NVIDIA Tesla T4 GPU for 250 epochs using stochastic gradient descent and the initial learning rate is set to be 0.01.
For SGConv, the kernel size of graph convolution is equal to the history horizon. The kernel size of each TGConv and temporal convolution is set to 3 and the kernel size used on VGG to extract the image features is $3\times 3$. The initial number of  input channel of SpecTGCNN unit is 2 and number of output channel is set to be 5. 
The optimal configuration was determined by validation, which is two layers of SpecTGNN units, two heads of STAtt and a set of temporal convolutions with five residual connections. 

\vspace{-0.1cm}

\subsection{Quantitative Analysis}

For the SDD dataset, we compared SpecTGNN with baselines in terms of minADE$_{20}$ and minFDE$_{20}$ in the pixel space. 
As shown in Table~\ref{scores_SDD}, SpecTGNN achieves the best results on both metrics. 
More specifically, our model achieves a performance improvement of 17.6\% on minADE$_{20}$ and 21.9\% on minFDE$_{20}$ compared to the previous state-of-the-art method.
Our model takes advantage of the spectral-temporal graph convolution and can jointly identify structural and sequential patterns in frequency domain, which can represent the correlation between agents more effectively than previous spatio-temporal graph convolutional network.
Compared with the S-STGCNN which employs the spatial and temporal convolution separately, results show that our design achieves better performance. 

For the nuScenes dataset, we compared SpecTGNN with the baselines with publicly available code and conducted all the experiments under the same settings. 
We predicted the trajectory at future 4s based on the previous 4s observations.
As shown in Table~\ref{scores_nu}, our model outperforms the baselines by a large margin, especially for long-term prediction. 
Compared with the previous state-of-the-art baselines STGAT~\cite{huang2019stgat} and S-STGCNN~\cite{mohamed2020social} which only leverage the spatial and temporal information, the proposed model consistently performs better, indicating that the information in frequency domain can further improve the conventional graph neural network. SpecSTGNN reduces  minFDE$_{20}$ of the final 4.0s by 8.1\% on the nuScenes dataset compared with the SOTA baseline STGAT.

\subsection{Ablation Analysis}
To better understand the significance of components in SpecTGNN, we conducted ablation experiments on the SDD dataset with four variants and on the nuScenes dataset with three variants.  
The results are summarized in the right parts of Table \ref{scores_SDD} and Table \ref{scores_nu}, which show that all the components are effective and indispensable. 
We elaborate each variant:
\begin{itemize}

\item \textbf{Base}: The base SpecTGNN only contains three modules: ``Agent Modeling'', ``A-SpecTGNN Block'' and ``TCNN''. Note that in order to illustrate the effectiveness of the proposed sequence operations in frequency domain, the TGConv module in A-SpecTGNN block is removed from the complete model.
 
\item \textbf{+TGConv}: The TGConv operation is added inside the SpecTGNN block to the base model.

\item \textbf{+STAtt}: The STAtt module is added to the base model, which can help strengthen the recognition of dependence between different prediction time steps over a long-term horizon.

\item \textbf{+Image}: This setting additionally equips the base model with image information of the environment.

\end{itemize}

Comparing \textbf{+TGConv} with the \textbf{Base} setting, we can see that \textbf{+TGConv} leads to performance improvement by about 9.0\% on minADE$_{20}$, since temporal dependency captured by TGConv provides important clues for time-series prediction. 
Compared with the \textbf{Base} model, the \textbf{+STAtt} setting reduces minADE$_{20}$ by 5.4\% and minFDE$_{20}$ by 3.9\% on the SDD dataset. 
Results show that \textbf{+Image} reduces minADE$_{20}$ by 11.2\%  and minFDE$_{20}$ by 9.6\% on the SDD dataset compared with \textbf{Base} model.
The complete model \textbf{SpecTGNN} preforms best in all experiments.

\subsection{Analysis on the number of samples $K$}
\begin{figure}
\centering
\includegraphics[width=\columnwidth,height=3.5cm]{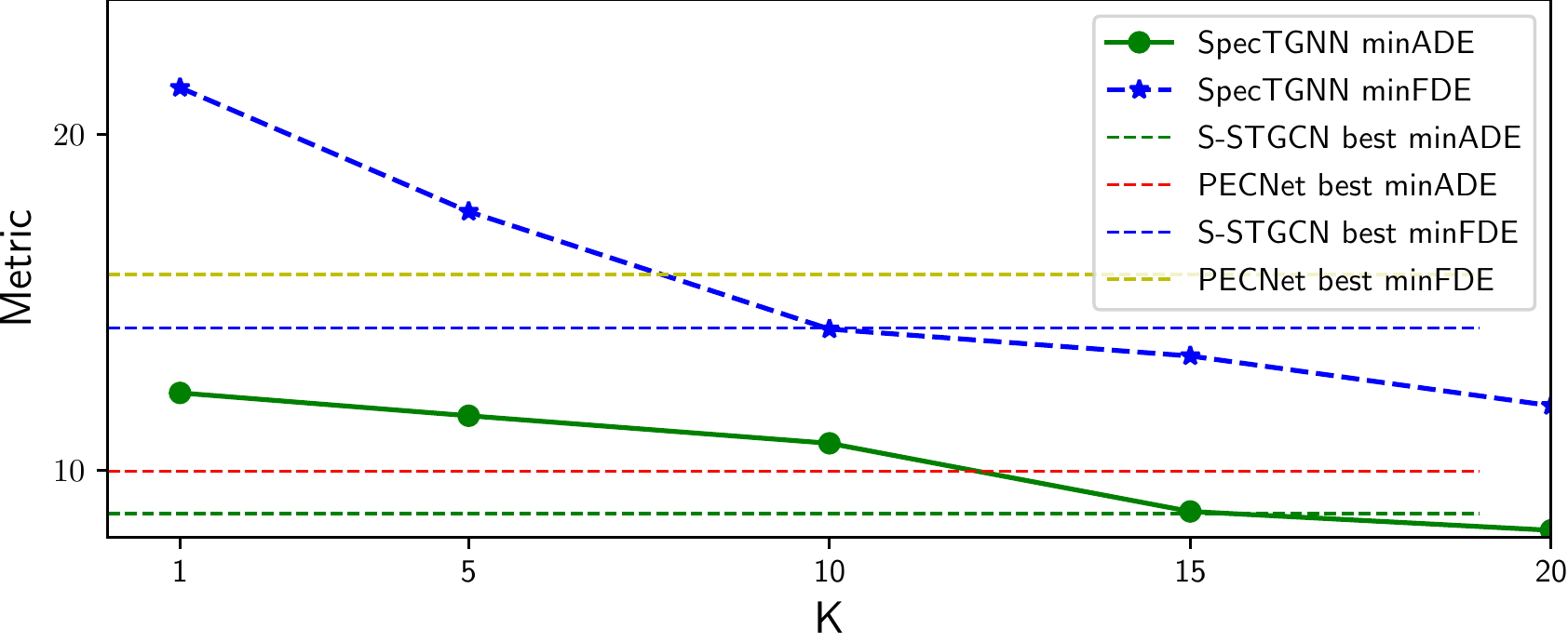}
\caption{Effect of $K$: the performance of minADE and minFDE against the number of samples used for evaluation.}
\vspace{-0.4cm}
\label{K}
\end{figure}

For the SDD dataset, we used $K = 20$ samples to evaluate the prediction accuracy in terms of minADE and minFDE, which is a widely used amount of hypotheses. In addition, we compare the results with different $K$ values. 
It can be seen from Fig. \ref{K} that as $K$ increases, both minADE$_K$ and minFDE$_K$ present a consistent decreasing trend. 
We can tell that our method can achieve comparable performance to the previous state-of-the-art baseline with much fewer samples. 
When fixing the number of samples, our method outperforms all baselines, which further supports our hypothesis that the modeling of structural and temporal relationship in frequency domain significantly reduces the prediction error.

\subsection{Qualitative Analysis}

\begin{figure}[!tbp]
    \centering
    \subfigure[]{
        \includegraphics[width=3.5cm, height=3.5cm]{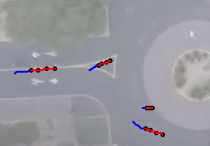}
        \label{label_for_cross_ref_1}
    }
    \subfigure[]{
        \includegraphics[width=3.5cm, height=3.5cm]{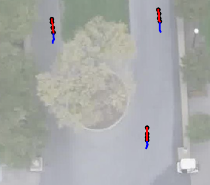}
        \label{label_for_cross_ref_3}
    }
    
    \subfigure[]{
	\includegraphics[width=3.5cm, height=3.5cm]{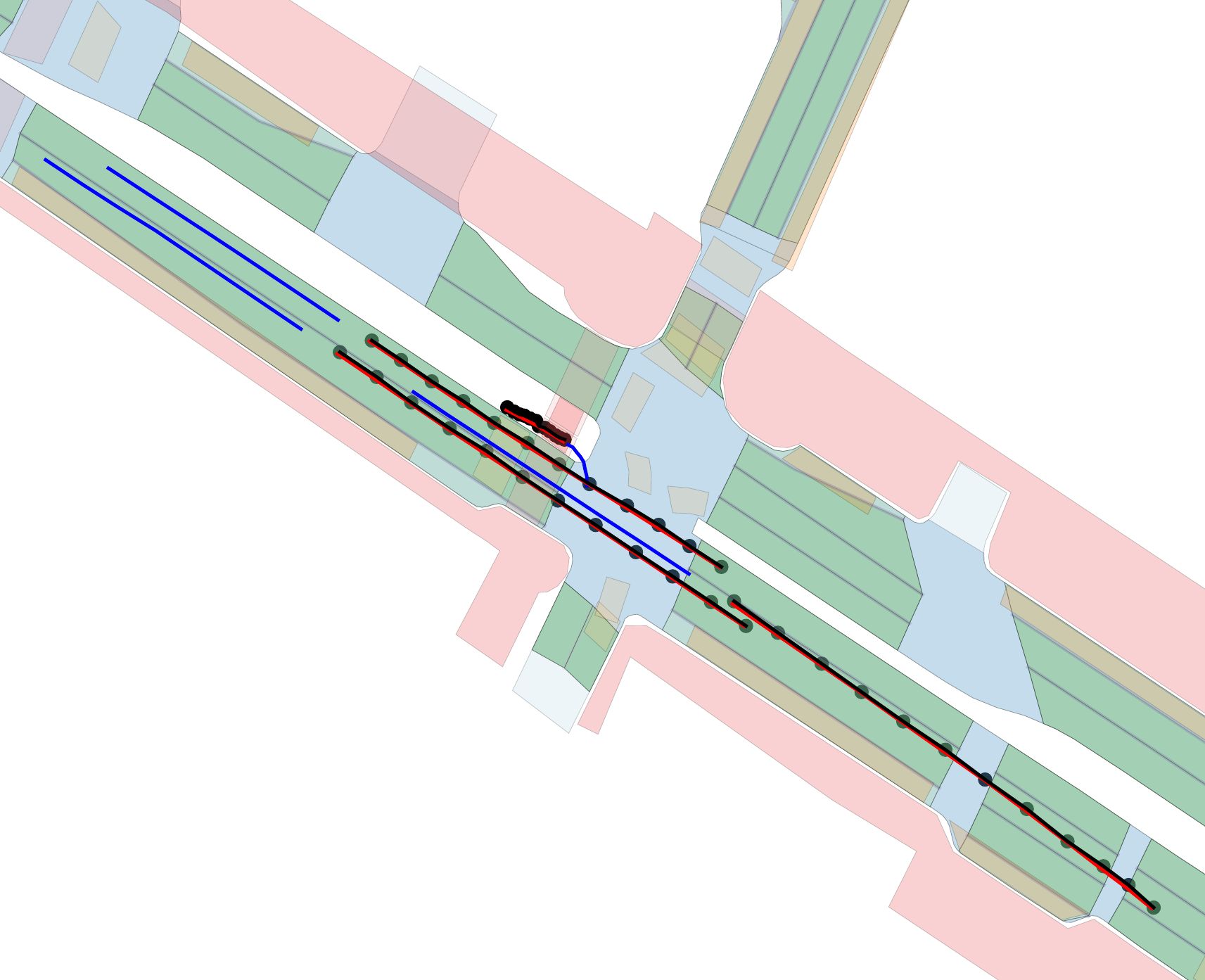}
        \label{label_for_cross_ref_2}
    }
    \subfigure[]{
	\includegraphics[width=3.5cm, height=3.5cm]{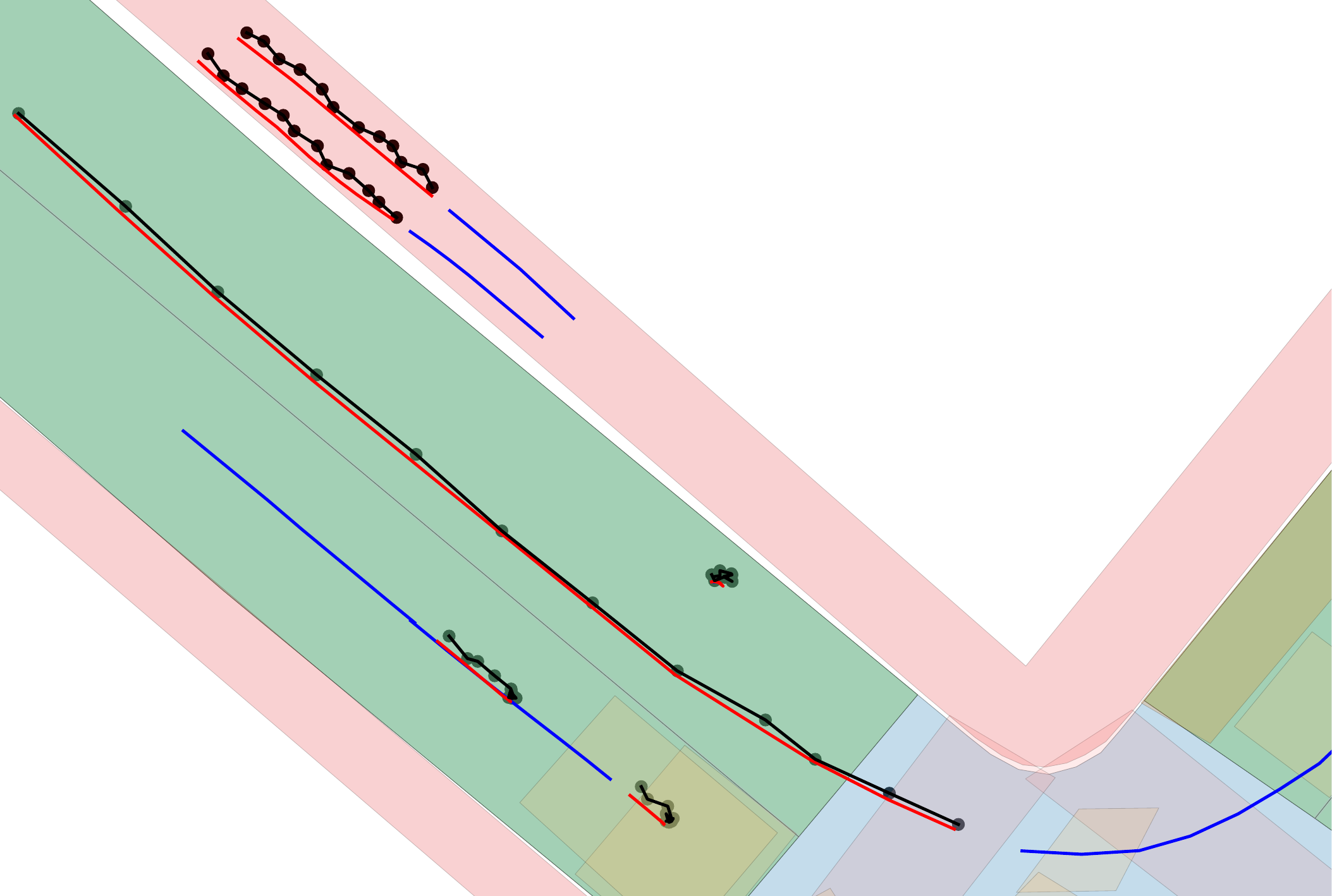}
        \label{label_for_cross_ref_4}
    }
    \caption{Qualitative results of the SDD dataset (a)(b) and nuScenes dataset (c)(d). The black dots represent the best predicted trajectory of each agent from 20 prediction hypotheses. The blue lines and red lines represent historical observation and ground truth, respectively.}
    \label{fig.1}
    \vspace{-0.5cm}
\end{figure}

We provide qualitative results of typical test cases in various traffic scenarios in Fig.~\ref{fig.1}.
The results show that our method is able to accurately predict future trajectories of the agents in the scene.
More specifically, as shown in Fig.~\ref{fig.1}(a), our model can handle roundabout scenes with different types of agent behaviors, including turning, stopping and going straight.
In Fig.~\ref{fig.1}(b), SpecTGNN can generate reasonable and accurate prediction hypotheses that are located in the feasible and traversable areas of pedestrians.
Fig.~\ref{fig.1}(c) shows that SpecTGNN is able to handle the opposite directions and the significant change in direction of movement. In Fig.~\ref{fig.1}(d), both the turning and lane keeping behavior can be accurately forecasted by our model.

\vspace{-0.1cm}
\section{Conclusion}
\vspace{-0.1cm}

In this paper, we propose a novel trajectory prediction model (SpecTGNN), which considers the interaction between agents and the influence of environment information simultaneously. 
The proposed SpecTGNN unit realizes sequence modeling and spatial graph convolution modeling of the trajectory jointly in frequency domain. 
In addition, the STAtt module can further enhance the exploitation of long-term temporal dependency of the embedded trajectory data.
The effectiveness of SpecTGNN is validated by pedestrian and vehicle trajectory prediction tasks on two benchmark datasets. 
Experimental results show that our method achieves state-of-the-art prediction performance compared with the baselines. In future work, we will explore more effective ways to build the graph structure and stricter distinctions will be made between pedestrians and vehicles.

\bibliographystyle{IEEEtran}
\bibliography{IEEEabrv,references}

\end{document}